\documentclass{llncs}

\pdfoutput=1

\usepackage{times}

\usepackage[utf8]{inputenc}
\usepackage{graphicx}
\usepackage{tikz}
\usepackage{pgfplots}
\usetikzlibrary{plotmarks}
\usepackage[english]{babel}
\usepackage{url}
\usepackage[ruled,vlined,linesnumbered]{algorithm2e}
\usepackage{algpseudocode}

\usepackage{amsmath, amssymb}
\usepackage{multirow}
\usepackage{amssymb}
\usepackage{colortbl,array}
\usepackage{graphicx}
\usepackage{mathtools}
\usepackage{color}
\usepackage{empheq}
\usepackage{subfig}
\usepackage{xspace}
\usepackage{xcolor}

\usepgfplotslibrary{units}

\newcolumntype{L}[1]{>{\raggedright\let\newline\\\arraybackslash\hspace{0pt}}m{#1}}
\newcolumntype{C}[1]{>{\centering\let\newline\\\arraybackslash\hspace{0pt}}m{#1}}
\newcolumntype{R}[1]{>{\raggedleft\let\newline\\\arraybackslash\hspace{0pt}}m{#1}}

\newcommand{\lang}[1]{\mathcal{L}_{#1}}

\def\N{\mbox{$\mathbb{N}$}}

\def\closed{\textsc{ClosedPattern}\xspace}
\def\closedx{\textsc{ClosedPattern}}
\def\cpim{\texttt{CP4IM}\xspace}

\def\pattern{pattern\xspace}
\def\patterns{patterns\xspace}
\def\card{\#}

\newcommand{\angx}[1]{{{\mbox{$\langle #1 \rangle$}}}}

\newcommand{\SDB}{\mathcal{D}}

\newcommand{\I}{\mathcal{I}}

\usepackage{etoolbox}
\newbool{seeAll}
\usepackage[normalem]{ulem}

\def\charm{\texttt{CHARM}\xspace}

\def\lcm{\texttt{LCM}\xspace}

\def\lcmthree{\texttt{LCM-v3}\xspace}
\def\closet{\texttt{CLOSET}\xspace}

\def\close{\texttt{CLOSE}\xspace}

\newcommand{\bdd}{\mathcal{T}}				
\newcommand{\items}{\mathcal{I}} 			

\newcommand{\freq}[0]{\ensuremath{f\!req}\xspace}

\begin{document}

\title{A global constraint for closed itemset mining}
\author{M. Maamar$^{a,b}$ \and N. Lazaar$^{b}$ \and S. Loudni$^{c}$ \and Y. Lebbah$^{a}$}
\institute{
		$(a)$ University of Oran 1, LITIO, Algeria. \\
		$(b)$ University of Montpellier, LIRMM, France. \\
		$(c)$ University of Caen, GREYC, France.\\
}

\maketitle

\vspace*{-.8cm}
\begin{abstract}
Discovering the set of closed frequent patterns is one of the fundamental problems in Data Mining. 
Recent Constraint Programming (CP) approaches for declarative itemset mining have proven their usefulness and flexibility. 
But the wide use of reified constraints in current CP approaches raises many difficulties to cope with high dimensional datasets.
This paper proposes \closed global constraint which does not require any reified constraints nor any extra variables to encode efficiently the Closed Frequent Pattern Mining (CFPM) constraint. \closed captures the particular semantics of the CFPM problem in order to ensure a polynomial pruning algorithm ensuring domain consistency.
The computational properties of our constraint are analyzed and their practical effectiveness is experimentally evaluated.



\end{abstract}

\section{Introduction}
\label{sec-introduction}

Frequent Pattern Mining is a well-known and the most popular research field of data mining. 
Originally introduced by~\cite{agrawal1993mining}, it plays a key role in many data mining applications. 
These applications include the discovery of frequent itemsets and association rules ~\cite{agrawal1993mining}, correlations~\cite{DBLP:conf/sigmod/BrinMS97} and many other data mining tasks.

In practice, the number of frequent patterns produced is often huge and can easily surpass the size of the input dataset. 
Based on this statement, it was important to identify a condensed representation of frequent patterns. 
On the other hand, most of frequent patterns are redundant where it is possible to derive them from other found patterns. 
That is, {\it closed frequent patterns} are one of the concise and condensed representations avoiding redundancy. 


Discovering the set of closed frequent patterns is one of the fundamental problems in Data Mining. 
Several specialized approaches have been proposed to discover closed frequent patterns (e.g., {\sc A-Close} algorithm ~\cite{DBLP:journals/is/PasquierBTL99}, \charm~\cite{DBLP:conf/sdm/ZakiH02}, \closet~\cite{DBLP:conf/dmkd/PeiHM00}, \lcm~\cite{DBLP:conf/dis/UnoAUA04}). 


Over the last decade, the use of Constraint Programming paradigm (CP) to model and to solve Data Mining problems has received a considerable attention~\cite{de2008constraint,guns2013k,DBLP:conf/cp/KhiariBC10}. 
The declarative aspect represents the key success of the proposed CP approaches. 
Doing so, one can add/remove any user-constraint without the need of developing specialized resolution methods.


Relating to the Closed Frequent Pattern Mining problem (CFPM), Luc De Raedt et. al., propose to express the different constraints that we can have in Pattern Mining as a CP model \cite{guns2013k}. 
The model is expressed on boolean variables representing items and transactions, with a set of reified sums as constraints.
The drawback is the wide use of reified constraints in the CP model, which makes the scalability of the approach questionable.

This paper proposes \closed global constraint which does not require any reified constraints nor any extra variables to encode efficiently the Closed Frequent Pattern Mining (CFPM) constraint. 
\closed captures the particular semantics of the CFPM problem in order to ensure a polynomial pruning algorithm ensuring domain consistency.

{Experiments on several known large datasets show that our approach clearly outperforms CP4IM~\cite{de2008constraint} and achieves scalability while it is a major issue for CP approaches. These experiments also show
that the fewer the number of closed patterns, the better is the performance of \closed. This is an
expected result of the fact that \closed insures domain consistency.


{
	The paper is organized as follows. Section~\ref{sec:cbl} recalls preliminaries. Section~\ref{lab-related} provides a critical review of specialized methods and CP approaches for CFPM. Section~\ref{sec-global4im} presents the global constraint \closed. Section~\ref{sec-experiments} reports experiments we performed. Finally, we conclude and draw some perspectives.}

\section{Background}
\label{sec:cbl}

\def\vide{\mbox{$\Box$}} 
\def\ID{\mbox{ID}} 

In this section, we introduce some useful notions in closed frequent pattern mining and constraint programming.
\subsection{Closed frequent \pattern mining}
Let $\items=\{1, ..., n\}$ be a set of $n$ \textit{items} identifiers and
$\bdd = \{1,...,m\}$ a set of \textit{transactions} identifiers. 
A \pattern $p$ (i.e., itemset) is a  subset of $\items$. 
The language of \patterns corresponds to $\lang{\items} = 2^{\items}$.  
A transaction database is a set $\SDB$ $\subseteq \items \times \bdd$. 
The set of items corresponding to a transaction identified by $t$ is denoted 
$\SDB[t]=\{i \,|\, (i, t)\in \SDB\}$.
A transaction $t$ is an occurrence of some pattern $p$ iff the set $\SDB[t]$ contains 
$p$ (i.e. $p \subseteq \SDB[t]$). 

The cover of $p$, denoted by $\bdd_{\SDB}(p)$, is the set of transactions containing 
$p$ (i.e. $\bdd_{\SDB}(p) = \{t \in \bdd \,|\, p \subseteq \SDB[t]\} $). 
Given $S\subseteq \mathcal{T}$ a subset of transactions, $\items_{\SDB}(S)=\bigcap_{t\in S}\SDB[t]$ is the set of common items of $S$. 
The (absolute) frequency of a pattern $p$ is the size of its 
cover (i.e., $\freq_{\SDB}(p)$ = $|\bdd_{\SDB}(p)|$). 
Let $\theta\in\N^+$ be some given constant called a {\it minimum support}. A pattern $p$ is frequent if $\freq_{\SDB}(p)\geq \theta$. 

\begin{example} Consider the transaction database in 
Table~\ref{table:transactionaldataset}. We 
have $\bdd_{\SDB}(CE) = \{2,3,5,6\}$
, $\freq_{\SDB}(CE) =4$
and 
$\items_{\SDB}(\{2,3,5,6\}) = BCE$.
\end{example}

The closure of a pattern $p$ in $\SDB$ is the set of common items of its cover $\bdd_{\SDB}(p)$, which is denoted $Clos(p)=\items_{\SDB}(\bdd_{\SDB}(p))$. A pattern is closed $closed_{\SDB}(p)$ 
 if and only if $Clos(p)=p$. 


\begin{definition}[Closed Frequent Pattern Mining (CFPM)]
Given a transaction database $\SDB$ and a minimum support threshold $\theta$. 
The closed frequent pattern mining problem is the problem of finding all patterns $p$ such 
that $(\freq_{\SDB}(p)\geq\theta)$ and $(Clos(p)=p)$.
\end{definition}

\begin{example} 
For $\theta = 2$, the set of closed frequent 
patterns in Table \ref{table:transactionaldataset} is $C \langle 5 \rangle$\footnote{Value between $\langle
. \rangle$ indicates the frequency of a pattern.},  
$BE \langle 5 \rangle, BCE\langle 4
\rangle, ABCE \langle 2\rangle$ and $AC \langle 2 \rangle$.
\end{example}


Closed frequent \patterns provide a minimal representation of frequent patterns
, i.e.,  we can derive all frequent patterns with their 
exact frequency value 
from the closed ones. 

\begin{table}[t]
\begin{minipage}{5cm}
\centering
\subfloat[ \label{table:transactionaldataset}]{
\scalebox{0.9}{
\begin{tabular}{c*5{@{\ }c@{\ }}}
\hline
t & \multicolumn{5}{c}{Items}        \\ 
\hline
$t_1$ & $A$ &     & $C$ & $D$ &     \\ 
$t_2$ &     & $B$ & $C$ &     & $E$ \\ 
$t_3$ & $A$ & $B$ & $C$ &     & $E$ \\ 
$t_4$ &     & $B$ &     &     & $E$ \\ 
$t_5$ & $A$ & $B$ & $C$ &     & $E$ \\ 
$t_6$ &     & $B$ & $C$ &     & $E$ \\ \hline

\end{tabular}}
}
\end{minipage}
\begin{minipage}{0.5cm}
\subfloat[\label{table2:binarymatrix}]{
\scalebox{0.9}{
\begin{tabular}{c*5{@{\ }c@{\ }}}
\hline
t & $A$ & $B$ & $C$ & $D$ & $E$\\ 
 \hline
$t_1$ & $1$ & $0$ & $1$ & $1$ & $0$ \\
$t_2$ & $0$ & $1$ & $1$ & $0$ & $1$ \\
$t_3$ & $1$ & $1$ & $1$ & $0$ & $1$ \\
$t_4$ & $0$ & $1$ & $0$ & $0$ & $1$ \\
$t_5$ & $1$ & $1$ & $1$ & $0$ & $1$ \\
$t_6$ & $0$ & $1$ & $1$ & $0$ & $1$ \\
\hline
\end{tabular}}
}
\end{minipage}
\caption{\small A transaction database $\SDB$ (a) and its binary matrix (b). }
\label{figure:runningexample}
\end{table}

\noindent
{\bf Search Space Issues.} 
In pattern mining, the search space contains $2^{\items}$  candidates. 
Given a  large number of items $\items$, a naïve search that consists of enumerating and testing the frequency of pattern candidates in a given dataset is infeasible. 
The main property exploited by most algorithms to reduce the search space is 
that frequency is {\it monotone decreasing} with respect to extension of a set.   

\begin{property}[Anti-monotonicity of the frequency] \label{antimonotone}
Given a transaction database $\SDB$ over $\items$, and two patterns  
$X$, $Y \subseteq \items$. Then, $X \subseteq Y \rightarrow 
\freq_{\SDB}(Y) \leq \freq_{\SDB}(X)$.
\end{property}

Hence, any subset (resp. superset) of a frequent (resp. infrequent)  pattern is also a frequent (resp. infrequent) pattern.

\subsection{CFPM under constraints}
Constraint-based pattern mining aims at extracting all patterns $p$ of 
$\lang{\items}$ satisfying a query $q(p)$ (conjunction of constraints), 
which usually defines what we call a \textit{theory}~\cite{MT97}: 
$Th(q) = \{p \in \lang{\items} \mid {q}(p) \,\, is \,\, true\}$. 
A common example is the frequency measure leading to the minimal frequency 
constraint.
It is also possible to have other kind of (user-)constraints. 
For instance,
Constraints on the size of the returned patterns,  $minSize(p,\ell_{min})$ constraint (resp. $maxSize(p,\ell_{min})$) holds if the number of items of $p$ is greater or equal (resp. less or equal) to $\ell_{min}$. 
Constraints on the presence of an item in a pattern $item(p,i)$ that states that an item $i$ must be (or not) in a pattern $p$. 
%
%
%
%
%


\subsection{Dataset representations}
The algorithms for frequent pattern mining differ mainly on the way that the dataset is represented.


{\it Horizontal representation $\mathcal{H}$.} Here, the transaction dataset is represented as a list of transaction. 
The Apriori algorithm~\cite{agrawal1993mining} is one of the approaches that adopt this obvious
representation. The drawback of such representation is the fact that we need several passes to update the support of pattern candidates.


{\it Vertical representation $\mathcal{V}$}. This representation uses a list of items where for each item, we have the list of transactions 
where it appears. Many algorithms adopt this representation~\cite{DBLP:conf/kdd/ZakiPOL97,DBLP:conf/kdd/ZakiG03,DBLP:BurdickCFGY05}. 
The key advantage of using such representation is that the support of a pattern candidate can easily be obtained by intersecting 
the lists of its items.

{\it Hybrid representation $\mathcal{HV}$.} The transaction dataset here is represented dually, horizontally and vertically.  It is successfully used within LCM \cite{DBLP:conf/dis/UnoAUA04} and FPgrowth \cite{DBLP:journals/datamine/HanPYM04} algorithms. 



\subsection{CSP and Global Constraints}

\noindent
A {\it Constraint Satisfaction Problem} (CSP) consists of a 
set $X$ of $n$ variables, a domain $\SDB$ mapping each variable 
$X_i \in X$ to a finite set of values $D(X_i)$, and a set of constraints
$\mathcal{C}$. An assignment $\sigma$ is a mapping from variables in $X$ to
values in their domains. 
A constraint $c \in \mathcal{C}$ is a subset of the 
cartesian product of the domains of the variables that are in $c$. 
The goal is to find an assignment 
such that all constraints are satisfied.  

\noindent 
\textbf{Domain consistency (DC).} 
Constraint solvers typically use backtracking search to explore the search
space of partial assignments. At each assignment, filtering
algorithms prune the search space by enforcing local consistency
properties like domain consistency. A constraint $c$ on $X$ is domain
consistent, if and only if, for every  $X_i \in X$ and every
$d_i \in D(X_i)$, there is an assignment $\sigma$ satisfying $c$ such
that $X_i = d_i$. 

\noindent 
\textbf{Global constraints} 
are constraints capturing a relation between a non-fixed number of variables.
These constraints provide the solver with a better view of the structure of the problem. Examples of global constraints are AllDifferent, Regular and Among (see \cite{Rossi06}). Global constraints cannot be efficiently propagated by generic local consistency algorithms, which are exponential in the number of the variables of the constraint. Dedicated filtering algorithms are constructed to achieve polynomial time complexity in the size of the input, i.e., the domains and extra parameters. This is the aim of this paper, which proposes a filtering algorithm for the frequent closed pattern constraint.

\section{Related works}
\label{lab-related}
This section provides a critical review of ad specialized methods and CP approaches for CFPM.

\smallskip
\noindent
{\bf Specialized methods for CFPM.}
\close~\cite{DBLP:journals/is/PasquierBTL99} was the first algorithm proposed to extract closed frequent patterns (CFPs). It uses an Apriori-like bottom-up method. Later, Zaki and Hsiao~\cite{DBLP:conf/sdm/ZakiH02} proposed a depth-first algorithm based on a vertical database format e.g. \charm. 
In~\cite{DBLP:conf/dmkd/PeiHM00}, Pei et al. extended the FP-growth method to a method 
called \closet for mining CFPs. 
Finally, Uno et al. \cite{DBLP:conf/dis/UnoAUA04} have proposed \lcm, one of the most fastest frequent itemset mining algorithm. It employs a hybrid representation based on vertical and horizontal ones. 

\smallskip
\noindent
{\bf CP methode for itemset mining.}
Luc De Raedt et al. have proposed in \cite{de2008constraint} a CP model for itemset mining (\cpim). They show how some constraints (e.g. frequency, maximality, closedness) can be modeled as CSP \cite{DBLP:conf/pkdd/NijssenG10,guns2011itemset}. This modeling uses two sets of boolean variables $M$ and $T$: (1) item variables $\{M_1, M_2,..., M_n\}$ where, given a pattern $P$, $(M_i = 1)$ iff
  $(i \in P)$; (2) transaction variables $\{T_1, T_2,..., T_m\}$ where $(T_t =1)$ iff $(P \subseteq t)$.

The relationship between $M$ and $T$ is modeled by reified constraints stating that, for each transaction $t$, $(T_t = 1)$ iff $M$ is a subset of $t$. A great consequence is that the encoding of the frequency measure is straightforward: $\freq_\mathcal{D} (M) = \sum_{t \in T} T_t$. 
But such an encoding has a major drawback since it requires $(m=\card\bdd)$ reified constraints to encode the whole database. This constitutes a strong limitation of the size of the databases that could be managed. 
 

We propose in the next section the \closed global constraint to encode both the minimum 
frequency constraint and the closedness constraint. This global constraint does not require any reified constraints nor any extra
 variables. 




\section{\closed global constraint for CFPM}
\label{sec-global4im}
\newcommand{\VDB}{\mathcal{V}}	

\def\pun{^{+}}
\def\pze{^{-}}
\def\pzu{^{*}}

This section presents the \closed global constraint for the CFPM problem.
\subsection{Consistency checking and filtering}
\label{filtering}
Let $P$ be the unknown pattern we are looking for. The unknown pattern $P$ is encoded 
with boolean item variables $\{P_1, ..., P_n\}$,
where $D(P_i)=\{0, 1\}$ and 
($P_i=1$) iff $i\in P$. 
Let $\sigma$ be a partial assignment of variables $P$. 
$\sigma$ can be partitioned into three distinct subsets: 
\begin{itemize}
\item present items $\sigma\pun=\{i\,  \in \items \,|\, D(P_i)=\{1\}\}$, 
\item absent items $\sigma\pze=\{i\, \in \items \,|\, D(P_i)=\{0\}\}$, 
\item free items $\sigma\pzu=\{i\, \in \items \,|\, D(P_i)=\{0,1\}\}$. 
\end{itemize}

The global constraint \closed ensures both 
minimum frequency constraint and closedness constraint.  

\begin{definition}[\closed global constraint]\label{def-freqclosed}
The \closedx$(P, \SDB, \theta)$  constraint holds if and only if  
there exists an assignment $\sigma=\angx{d_1, ..., d_{n}}$ of
variables $P$ such that $\freq_{\SDB}(\sigma\pun)\geq\theta$ and $closed_{\SDB}(\sigma\pun)$. 
\end{definition}

\begin{example} 
\label{ex4}
	Consider the transaction database of Table \ref{table:transactionaldataset} with $\theta = 2$.
	Let $P = \angx{P_{1}, \dots, P_{5}}$ with $D(P_i) = \{0,1\}$ for $i=1..5$. 
	Consider the closed pattern $BCE$ encoded by $P=\angx{01101}$, where 
	$\sigma\pun=\{B,C,E\}$ and $\sigma\pze=\{A,D\}$.
	\closedx$(P, \SDB, 2)$ holds since $\freq_{\SDB}(\{B,C,E\})\geq 2$ and 
	$closed_{\SDB}(\{B,C,E\})$. 	
\end{example}


\begin{definition}[Full extension item \cite{DBLP:journals/widm/Borgelt12}] \label{full}
Let $\sigma =\angx{d_{i_1}, \dots, d_{i_\ell}}$ be a partial assignment of $\ell$ variables 
$\angx{P_{i_1}, \dots, P_{i_\ell}}$, and 
	$j$ an item such that  $j\not\in \sigma\pun$. 
	The item $j$  is called a full extension of $\sigma$ iff 
	$\bdd_{\SDB}(\sigma\pun) = \bdd_{\SDB}(\sigma\pun\cup \{j\})$. 
\end{definition}


Let $\sigma$ be a partial assignment of variables $P$ and $i$ a free item. 
We denote by $\VDB_{\SDB}^{\sigma\pun}(i)$ the cover of item $i$ within the current cover of a pattern 
$\sigma\pun$: 
\[\VDB_{\SDB}^{\sigma\pun}(i) = \bdd_{\SDB}(\sigma\pun\cup\{i\}) = \bdd_{\SDB}(\sigma\pun) 
\cap\bdd_{\SDB}(\{i\}).\] 

We first show when a partial assignment is consistent with respect to \closed constraint.

\begin{proposition}
\label{prop-full-item}
Let $\sigma =\angx{d_{i_1}, \dots, d_{i_l}}$ be a partial assignment of $\ell$ variables 
$\angx{P_{i_1}, \dots, P_{i_\ell}}$. We say that $\sigma$ is a consistent partial assignment iff:
$\freq_{\SDB}(\sigma\pun)\geq \theta $ and  $\not \exists j \in \sigma\pze$ s.t. $j$ is a full 
extension of $\sigma$.

\end{proposition}

\noindent
{\it Proof: }
\begin{description}
\item According to the anti-monotonicity property of the frequency (cf. property \ref{antimonotone}), if the partial assignment $\sigma$ is infrequent (i.e., $\freq_{\SDB}(\sigma\pun)< \theta$), it cannot, under any circumstances, be extended to a closed pattern.

\item Given now a frequent partial assignment $\sigma$ (i.e., $\freq_{\SDB}(\sigma\pun)\geq \theta$), let us take $j\in\sigma\pze$ s.t. $j$ is a full extension of $\sigma$. It follows that $\bdd_{\SDB}(\sigma\pun) = \bdd_{\SDB}(\sigma\pun \cup \{j\}) = \VDB_{\SDB}^{\sigma\pun}(j)$. Therefore, $Clos(\sigma\pun)=Clos(\sigma\pun \cup \{j\})$. Since $\sigma\pun$ without $j$ ($j$ being in $\sigma\pze$) cannot be extended to a closed pattern, the result follows. {If there is no item $j\in\sigma\pze$ s.t. $j$ is a full extension of $\sigma$, then the current assignment $\sigma$ can be definitely extended to a closed itemset by adopting all the full extension items to form a closed pattern. $\Box$}

\end{description}


We now give the \closed filtering rules by showing when a value of a given variable is inconsistent.
\begin{proposition}[\closed Filtering rules]
\label{prop-consistency}
Let $\sigma =\angx{d_{i_1}, \dots, d_{i_\ell}}$ be a consistent partial
assignment of $\ell$
variables $\angx{P_{i_1}, \dots, P_{i_\ell}}$, and $P_{j}$ ($j\in \sigma\pzu$) be a free variable.
The following two cases characterize the inconsistency of the values $0$ and $1$ of $P_j$: 
\begin{itemize}
\item $0\not\in D(P_j)$ iff: $j$ is a full extension of $\sigma.\qquad\quad\ $(rule 1)
\item $1\not\in D(P_j)$ iff: 
$\left \{
\begin{array}{l l}  
  |\VDB_{\SDB}^{\sigma^+}(j)|< \theta \ \ \ \ \vee & \text{(rule 2)}\\
 \exists k\in \sigma^-,  \VDB_{\SDB}^{\sigma^+}(j)\subseteq \VDB_{\SDB}^{\sigma^+}(k). & \text{(rule 3)}\\
\end{array}
\right.$
\end{itemize}
\end{proposition}

\noindent
{\it Proof: }

Let $\sigma$ be a consistent partial assignment and $P_j$ be a free variable.
\begin{description}
\item [$0\not\in D(P_j):$] 

($\Rightarrow$) Let $0$ be an inconsistent value. In this case, $P_j$ can only take value 1. It means that $Clos(\sigma\pun)=Clos(\sigma\pun\cup \{j\})$. Thus, $\bdd_{\SDB}(\sigma\pun) = \bdd_{\SDB}(\sigma\pun \cup \{j\})$. By definition \ref{full}, $j$ is a full extension of $\sigma$.

($\Leftarrow$) Let $j$ be a full extension of $\sigma$, which means that
$Clos(\sigma\pun)=Clos(\sigma\pun\cup \{j\})$ (def. \ref{full}). The value $0$ is inconsistent where $j$ cannot be in $\sigma\pze$ (proposition \ref{prop-full-item}).       

\item [$1\not \in D(P_j):$] 

($\Rightarrow$) Let $1$ be an inconsistent value. This can be the case if the frequency of the current pattern $\sigma\pun$ is set up bellow the threshold $\theta$ by adding the item $j$ (i.e., $|\VDB_{\SDB}^{\sigma^+}(j)|< \theta$). {Or, $\sigma\pun\cup \{j\}$ cannot be extended to a closed itemset: this is the case when }it exists an item $k$ such that at each time the item $j$ belongs to a transaction in the database, $k$ belongs as well ($\VDB_{\SDB}^{\sigma^+}(j)\subseteq \VDB_{\SDB}^{\sigma^+}(k))$. Conversly, the lack of $k$ (i.e., $k\in\sigma\pze$) implies the lack of $j$ as well. This means that:  ($P_k=0 \Rightarrow P_j=0$).

($\Leftarrow$) {This is a direct consequence of proposition \ref{prop-full-item}. $\Box$}
\end{description}

\begin{example} 
	In line of example \ref{ex4}, consider a partial assignment $\sigma$ s.t. $\sigma\pun=\{B\}$  
	and $\sigma\pze=\emptyset$. \closedx$(P, \SDB, \theta)$ will remove value $1$ from 
	$D(P_4)$ (item $D$) and value $0$ from $D(P_5)$ (item $E$) since, resp., $|\VDB_{\SDB}^{\sigma^+}(D)|< 2$ 
	and $E$ is a full extension of $\sigma$. Now, we have $\sigma\pun=\{B,E\}$ and 
	$\sigma\pze=\{D\}$. Suppose that the variable $P_3$ is set to $0$ (item $C$). 
	Again, \closedx$(P, \SDB, \theta)$ will remove value $1$ from $D(P_1)$ (item $A$) since 
	the lack of $C$ implies the lack of $A$ in $\SDB$ (i.e., $\VDB_{\SDB}^{\sigma^+}(A) \subseteq \VDB_{\SDB}^{\sigma^+}(C)$). 
\end{example}




\subsection{\closed Filtering Algorithm}
\label{sec-filtering}

In this section, we present an algorithm enforcing domain consistency for \closed constraint. Algorithm \ref{AC4IM} maintains the consistency based on the specificity of the CFPM problem (see proposition \ref{prop-full-item} and \ref{prop-consistency}). \closed is considered as a global constraint since all variables share the same internal data structures that awake and drive the filtering.  

{\sc Filter-}\closed algorithm exploits internal data structures enabling to enhance the filtering process. 
At each call, we maintain incrementaly $\sigma=<\sigma\pun,\sigma\pze,\sigma\pzu>$  and the cover of $\sigma\pun$ (i.e., $ \bdd_{\SDB}(\sigma\pun)$). 
Using these two structures, one can check if an item is present or not in the vertical dataset $\mathcal{V}_\mathcal{D}$.

Algo.\ref{AC4IM} takes as input the vertical dataset $\mathcal{V}_\mathcal{D}$, a minimum support threshold $\theta$, the item $k$ of the last assigned variable $P_k$, the current partial assignment $\sigma$ where $\sigma\pzu\neq \emptyset$, and the variables $P$. 
As output, algo.\ref{AC4IM} will reduce de domain of $P_i$ and therefore, increase $\sigma\pun$ and/or $\sigma\pze$, and decrease $\sigma\pzu$.
The algorithm starts by checking if the last variable $P_k$ is instantiated to $0$ (line \ref{algo:zero}). 
In such case, the item $k$ is added to (resp. removed from) $\sigma\pze$ (resp. $\sigma\pzu$) (line \ref{algo:sigma1}).
Afterwards, we check if the lack of the item $k$ induces the lack of other free items of $\sigma\pzu$ (lines \ref{algo:rule3-a1}-\ref{algo:rule3-a2}), which corresponds to rule 3 of proposition \ref{prop-consistency}.
Otherwise, $P_k$ is assigned to 1 or we are in the level of the first call of the {\sc Filter-}\closed where no variable is instantiated. Here, we add to (resp. remove from) $\sigma\pun$ (resp. $\sigma\pzu$) the item $k$ if $P_k=1$ (line \ref{algo:un}). 
Now, the first thing to check is to verify if the current assignement is consistent or not (proposition \ref{prop-full-item}). The line \ref{algo:cover} is doing so by  checking if the frequency of the current assignment (i.e., the size of the cover  $\bdd_{\SDB}(\sigma^+)$) is below the threshold $\theta$.

The lines from \ref{algo:rule1a} to \ref{algo:rule3b} represents the application of proposition \ref{prop-consistency} on the remaining free items $\sigma\pzu$. 
First, lines \ref{algo:rule1aa} to \ref{algo:rule1b} prune the value $0$ from each item that is a full extension of the current assignment (rule 1 of prop.\ref{prop-consistency}). 
Second, lines \ref{algo:rule2a} to \ref{algo:rule2b} prune the value $1$ from the infrequent items (rule 2 of prop.\ref{prop-consistency}).
Finaly, lines  \ref{algo:rule3a} to \ref{algo:rule3b} implement the rule 3 of prop.\ref{prop-consistency} and prune the value $1$ from each free item $i\in \sigma\pzu$ s.t. its cover  is a superset of the cover of an absent item $j\in\sigma\pze$ ($\VDB_{\SDB}^{\sigma\pun}(i)\subseteq \VDB_{\SDB}^{\sigma\pun}(j)$).


\begin{algorithm}[t] \footnotesize
	\caption{{\sc Filter-}\closed ($\mathcal{V}_\mathcal{D},\theta,k,\sigma,P$) }\label{AC4IM} 
	\SetKw{Input}{Input}
	\SetKw{InOut}{InOut}
	
	\SetKw{In}{In}
	\SetKw{Out}{Out}
	\SetKw{InOut}{InOut}
	
	\SetKw{Output}{Output}
	\SetKw{return}{return}
	
	\DontPrintSemicolon

	\BlankLine
	\LinesNotNumbered
		
	\Input: 
	$\mathcal{V}_\mathcal{D}$ : vertical database; 	
	$\theta: $ minimum support; 
	$k:$ instantiated item\\
	\InOut:
	$P=\{P_1\ldots P_n\}$: boolean item variables;
	$\sigma:$ current assignment.	
	\BlankLine

\textbf{begin}
		
\label{algo:zero}		\If{$P_k = 0$}{

\label{algo:sigma1}				$\sigma^-\gets \sigma^- \cup \{k\}$; $\quad$
				$\sigma^*\gets \sigma^* \setminus \{k\}$\\
			
	\ForEach{$i\in \sigma^*:  \VDB_{\SDB}^{\sigma\pun}(i) \subseteq \VDB_{\SDB}^{\sigma\pun}(k)$}
			{
\label{algo:rule3-a1}				$D(P_i) \gets D(P_i) - \{1\}$; $\quad$\\
				$\sigma^-\gets \sigma^- \cup \{i\}$; $\quad$
\label{algo:rule3-a2}	$\sigma^*\gets \sigma^* \setminus \{i\}$\\
			}
			
		}
		\Else{				
		\lIf{$P_k=1$}{	
\label{algo:un}			$\sigma^+\gets \sigma^+ \cup \{k\}; \quad \sigma^*\gets \sigma^* \setminus \{k\}$
			}
	\lIf{
	\label{algo:cover}	$( \ |\bdd_{\SDB}(\sigma^+)| <  \theta)$}{\return $false$}

			\ForEach{$i \in \sigma^*$}
			{

\label{algo:rule1a}				\If{$|\VDB_{\SDB}^{\sigma\pun}(i)|=|\bdd_{\SDB}(\sigma^+)|$}{

\label{algo:rule1aa}				$D(P_i)\gets D(P_i)-\{0\}$; $\quad$ \\
				$\sigma^+\gets \sigma^+ \cup \{i\}$; $\quad$
\label{algo:rule1b}				$\sigma^*\gets \sigma^* \setminus \{i\}$

				}
			\ElseIf{$|\VDB_{\SDB}^{\sigma\pun}(i)|<\theta$ }{
\label{algo:rule2a}							$D(P_i) \gets D(P_i) - \{1\}$ \\
							$\sigma^-\gets \sigma^- \cup \{i\}$; $\quad$
\label{algo:rule2b}							$\sigma^*\gets \sigma^* \setminus \{i\}$\\
							}		
				\Else{
\label{algo:rule3a}					\ForEach{$j\in \sigma^- :  \VDB_{\SDB}^{\sigma\pun}(i) \subseteq \VDB_{\SDB}^{\sigma\pun}(j)$}
					{
						$D(P_i) \gets D(P_i) - \{1\}$; $\quad$\\
						$\sigma^-\gets \sigma^- \cup \{i\}$; $\quad$
\label{algo:rule3b}						$\sigma^*\gets \sigma^* \setminus \{i\}$\\
					}
				}
					
			}

	}

		\return $true$;

\end{algorithm}

\begin{theorem}
Given a transaction database $\mathcal{D}$ of $n$ items and $m$ transactions, and a threshold minsup $\theta$. Algorithm {\sc Filter-}\closed establishes domain consistency on the \closed constraint, or proves that it is  inconsitent in $O(n^2\times m)$ with a space complexity of {$O(n\times m)$}.  
\end{theorem}

\noindent
{\it Proof: }

\begin{description}
\item { Since that \closed implements exactly the rules given in Proposition \ref{prop-consistency}, it prunes all inconsistent values, and consequently ensures domain consistency (see the given description of Algorithm \ref{AC4IM}).}

\item Let $n=|\mathcal{I}|$ and $m=|\mathcal{T}|$. {First, we need to compute $\bdd_{\SDB}(\sigma^+)$ which requires at most $O(n\times m)$. This is done only once. The cover $\VDB_{\SDB}^{\sigma\pun}(i)$ can be computed by intersecting $\bdd_{\SDB}(\sigma^+)$ (already computed) and $\bdd_{\SDB}(\{i\})$ (given by the vertical representation) within at most $O(m)$.} Checking rule 1 and 2 on all free variables {can be done in $O(n\times m)$} ( lines \ref{algo:rule1aa}-\ref{algo:rule2b}). However, checking rule 3 is quadratic at lines \ref{algo:rule3-a1}-\ref{algo:rule3-a2} (i.e., $O(n\times m)$) and cubic at lines \ref{algo:rule3a}-\ref{algo:rule3b}  (i.e., $O(n\times(n\times m))$), where checking if a cover $\VDB_{\SDB}^{\sigma\pun}(i)$ is a subset of another cover can be done in $O(m)$. {Finally, the worst case complexity is $O(n\times(n\times m))$.}

\item The space complexity of the filtering algorithm lies in the storage of {$\mathcal{V}_\mathcal{D}$,} $\sigma$ and the cover $\mathcal{T}$ data structures. {The vertical representation $\mathcal{V}_\mathcal{D}$ requires at most $n\times m$. In the worst case, we have to store $n$ items within $\sigma$ and $m$ transactions within $\mathcal{T}$. That is, the worst case space complexity is $O(n\times m + n + m)$ = $O(n\times m)$. $\Box$}  	
\end{description}	
	
{During the solving process in depth first search, the whole space complexity is about $O(n\times (m + n))$ since that: (1) the depth is at most $n$; (2) $\sigma$ and $\mathcal{T}$ require $O(n\times (m + n))$; (3) the vertical representation is the same data used all along the solving process $O(n\times m)$; (4) Finally we have $O(n\times (m + n)) + O(n\times m) = O(n\times (m + n))$.}

\section{Experiments}
\label{sec-experiments}
We made some experiments to evaluate and compare our global constraint \closed with the state of the art methods (CP and specialized methods for CFPM).
We first present the benchmark datasets we used for our experiments and we give a brief description on the followed protocol.

{\bf Benchmark datasets.}
The reported results are on several real and synthetic 
datasets~\cite{DBLP:conf/sdm/ZakiH02,DBLP:journals/tkde/GrahneZ05} 
from FIMI repository\footnote{\url{http://fimi.ua.ac.be/data/}} with 
large size. 
These datasets have varied characteristics and representing different 
application domains as shown in Table~\ref{table:CharData}. 

Table~\ref{table:CharData} reports for each dataset, the number of transaction $|\mathcal{T}|$, the number of items $|\I|$, the average size of transaction $\widehat{|\mathcal{T}|}$ and its density $\rho$ (i.e., $\widehat{|\mathcal{T}|}/|\I|$).
As we can see, we select datasets by varying the number of transactions, the number of items, but also the density $\rho$. Here we have datasets that are very dense like chess and connect (resp. 49\% and 33\%), others that are very sparse like Retail and BMS-Web-View1 (resp. 0.06\% and 0.5\%).

\begin{table}[t] \centering
\scalebox{0.72}{
\begin{tabular}{|l|c|c|c|c|c|c|}
\hline
Dataset & $|\mathcal{T}|$ & $|\I|$ & { $\widehat{|\mathcal{T}|}$} & $\rho$  & type of data\\
\hline
Chess & 3 196 & 75  &37& 49\% & game steps\\
\hline
Connect & 67 557 & 129 &43& 33\% & game steps\\ 
\hline
Mushroom & 8 124 & 119  &23& 19\% & species of mushrooms \\
\hline
Pumsb & 49 046 & 7 117  &74& 1\%& census data \\
\hline
BMS-Web-View1 & 59 601 &  497  & 2.5& 0.5\%& web click stream \\
\hline
T10I4D100K & 100 000 & 1 000  &10& 1\%& synthetic dataset\\
\hline
T40I10D100K & 100 000 & 1 000 &40&4\% & synthetic dataset\\
\hline
Retail & 88 162 & 16 470  &10& 0.06\%& retail market basket data\\
\hline
\end{tabular}
}
\vspace*{-.15cm}
\caption{Dataset Characteristics.} 
\label{table:CharData}
\end{table}

\noindent
{\bf Experimental protocol.}
The implementation of our approach was carried out in the {\tt Gecode}
solver\footnote{\url{http://www.gecode.org}}. 
All experiments were conducted on an Intel Core i5-2400 @ 3.10 GHz with 8 Gb of RAM with 
a timeout of $3600s$. 
For each dataset, we varied the $minsup$ threshold until the
methods are not able to complete the extraction of all closed patterns
within the timeout limit. 
We compare our approach (\closed) with:
\begin{enumerate} 
\item \cpim, the most popular CP approach for CFPM.  
\item \lcm, the most popular specialized method for CFPM. 
\end{enumerate}

We experiment using the available distributions of  \lcmthree~\footnote{\url{http://research.nii.ac.jp/~uno/codes.htm}} and \cpim~\footnote{\url{https://dtai.cs.kuleuven.be/CP4IM/}}. We denote the fact that the underlying solver used by \cpim is {\tt Gecode} solver. 


\begin{figure*}[!ht]
\captionsetup[subfigure]{labelformat=empty}

\begin{center}
\scalebox{0.55}{
\begin{tabular}{|l|l|l|r|r|r|r|r|r|}
\hline
\multirow{2}{*}{Dataset} & \multirow{2}{*}{$minsup$ (\%)} &
\multirow{2}{*}{\#Pat} & \multicolumn{2}{c|}{\#Propagations} & \multicolumn{2}{c|}{\#Nodes} & \multicolumn{2}{c|}{Memory}\\
\cline{4-9} 
&  &  & \closed & \cpim & \closed & \cpim & \closed & \cpim \\
\hline
\multirow{8}{*}{Mushroom} 
&30 & 428 & {\bf 1447} & 430016 & {\bf 995} & 1039 & {\bf 18416} & 88079 \\
&20 & 1198 & {\bf 4250} & 1030445 & {\bf 2881} & 3071 & {\bf 31112} & 90831 \\
&10 & 4898 & {\bf 17221} & 2771719 & {\bf 11443} & 13281 & {\bf 42840} & 93839 \\
&5 & 12855 & {\bf 45361} & 5574143 & {\bf 30237} & 36495 & {\bf 69216} & 97552 \\
&1 & 51672 & {\bf 169781} & 13813312 & {\bf 117851} & 168999 & {\bf 100528} & 102608 \\
&0.5 & 76199 & {\bf 240618} & 18018929 & {\bf 170091} & 259427 & 111296 & {\bf 104017} \\
&0.1 & 164118 & {\bf 479630} & 31222435 & {\bf 350491} & 529289 & 143664 & {\bf 106577} \\
&0.05 & 203882 & {\bf 580333} & 36520438 & {\bf 430799} & 622145 & 148120 & {\bf 107025} \\
\hline
\multirow{5}{*}{Chess} 
&60 & 98393 & {\bf 314182} & 2661395 & 197063 & {\bf 196787} & 29712 & {\bf 15623} \\
&50 & 369451 & {\bf 1206361}& 9201740 & 751733 & {\bf 738907} & 31952 & {\bf 16520} \\
&40 & 1366834 & {\bf 4590519} & 30541475 & 2863847 & {\bf 2733735} & 39720 & {\bf 18568} \\
&30 & 5316468 & - & {\bf 104618207} & - & {\bf 10635019} & - & {\bf 19976} \\
&20 & 22918586 & - & {\bf 385399747} & - & {\bf 45901933} & - & {\bf 19847} \\ 
\hline
\multirow{6}{*}{Connect} 
&90 & 3487 & {\bf 12582} & 1865236 & 8677 & {\bf 6973} & {\bf 18320} & 561872 \\ 
&80 & 15108 &{\bf 59708} & 9453279 & 41939 & {\bf 30215} & {\bf 21752} & 585040 \\
&70 & 35876 & {\bf 147814} & 24968701 & 105663 & {\bf 71751} & {\bf 22848} & 594640 \\
&60 & 68350 & {\bf 281560} & 51648114 & 203425 & {\bf 136699} & {\bf 24896} & 661583 \\
&50 & 130102 & {\bf 538749} & 98600221 & 382441 & {\bf 260203} & {\bf 26968} & 665871 \\
\hline
\multirow{5}{*}{Pumsb} 
&95 & 111 & {\bf 371}  & - & {\bf 273} & - & {\bf 13600} & - \\ 
&90 & 1467 & {\bf 4760} & - & {\bf 3001} & - & {\bf 18416} & - \\
&85 & 8514 & {\bf 30624} & - & {\bf 19581} & - & {\bf 20920} & - \\
&80 & 33296 & {\bf 137104} & - & {\bf 89227} & - & {\bf 23280} & - \\
&75 & 101048 & {\bf 432278} & - & {\bf 290553} & - & {\bf 26688} & - \\
\hline
\multirow{6}{*}{Retail} 
&10 & 10 & {\bf 27} & - & {\bf 19} & - & {\bf 8760} & - \\
&5 & 17 & {\bf 48} & - & {\bf 33} & - & {\bf 11096} & - \\
&1 & 160 & {\bf 526} & - & {\bf 319} & - & {\bf 62568} & - \\
&0.5 & 581 & {\bf 1942} & - & {\bf 1161} & - & {\bf 425608} & - \\
&0.1 & 7696 & {\bf 25785} & - &  {\bf 15391} & - & {\bf 44152888} & - \\ 
&0.05 & 19699 & {\bf 65781}  & - & {\bf 39407} & - & {\bf 113664128} & - \\
\hline
\multirow{5}{*}{T10I4D100K} 
&1 & 386 & {\bf 1288} & - & {\bf 771} & - & {\bf 1207720} & - \\
&0.5 & 1074 & {\bf 3310} & - & {\bf 2147} & - & {\bf 3118024} & - \\
&0.1 & 26807 & {\bf 72073} & - & {\bf 53765} & - & {\bf 5074328} & - \\
&0.05 & 46994 & {\bf 132332} & - &  {\bf 95223} & - & {\bf 5519552} & - \\ 
&0.01 & 283398 & {\bf 935190}  & - & {\bf 602235} & - & {\bf 5947168} & - \\
\hline
\multirow{2}{*}{T40I10D100K} 
&10 & 83 & {\bf 276} & - & {\bf 165} & - & {\bf 143552} & - \\
&1 & 65237 & {\bf 176994} & - & {\bf 130473} & - & {\bf 4678544} & - \\
\hline
\multirow{5}{*}{BMS-Web-View1} 
&0.16 & 32 & {\bf 102} & - & {\bf 63} & - & {\bf 17464} & - \\
&0.08 & 9392 & {\bf 29854} & - & {\bf 18935} & - & {\bf 1112024} & - \\
&0.06 & 64763 & {\bf 220487} & - & {\bf 147811} & - & {\bf 1174472} & - \\
&0.04 & 155652 & {\bf 611247} & - & {\bf 424203} & - & {\bf 1253304} & - \\
&0.02 & 422693 & {\bf 1533714} & - & {\bf 1031727} & - & {\bf 1489272} & - \\ 
\hline
\end{tabular}
}
\end{center}

\centering
\subfloat[][]{
\begin{tikzpicture}[scale=0.46]
\begin{axis}[
             ymax = 50, ymin = 0.01,
             title={Mushroom},
             every axis title/.style={below right,at={(0.35,1.30)}},
             yscale = 0.89,
             ylabel={CPU time(s)},
             xlabel={minsup(\%)},
       		 symbolic x coords={$30$,$20$,$10$,$5$,$1$,$0.5$,$0.1$,$0.05$},
             xtick=data,
             nodes near coords align={vertical},
             legend entries={\closed, \cpim, \lcmthree},
             legend style={at={(0.10,1.00)},anchor=north west,{draw=none},font=\tiny},	
			 legend columns=1
             ]
    
\addplot+[mark=+,mark options = {scale=0.8},blue] coordinates {($30$,0.26)($20$,0.52) ($10$,1.30) ($5$,2.75) ($1$,6.29) ($0.5$,7.72) ($0.1$,10.60) ($0.05$,11.55) };
             
\addplot+[mark=*,mark options = {scale=0.8},red] coordinates {($30$,0.73)($20$,2.08) ($10$,5.17) ($5$,11.32) ($1$,26.22) ($0.5$,32.43) ($0.1$,45.19) ($0.05$,47.59)};

\addplot+[mark=diamond*,mark options = {scale=1.2},black] coordinates {($30$,0.008)($20$,0.010) ($10$,0.027) ($5$,0.030) ($1$,0.056)
($0.5$,0.29) ($0.1$,0.36) ($0.05$,0.42)};
\end{axis}
\end{tikzpicture}
}
\subfloat[][]{
\begin{tikzpicture}[scale=0.46]
\begin{semilogyaxis}[
					minor ytick={0.1,1,10,100,1000,10000},
					extra y ticks={3600},
					extra y tick style={grid=major},
					extra y tick labels={\color{black}TimeOut},
					yscale = 0.92,
					xmin = 1,
					x dir = reverse,
					every axis title/.style={below right,at={(0.42,1.20)}},
					title={Chess},
					ylabel={CPU time(s)},
             		xlabel={minsup(\%)},
             		legend entries={\closed, \cpim, \lcmthree},
             		legend style={at={(0.53,0.30)},anchor=north west,{draw=none},font=\tiny},	
					legend columns=1
					]
\addplot+[mark=+,mark options = {scale=0.8},blue] coordinates {
(60, 14.14) (50, 51.71) (40, 182.33) (30, 305.95) (20, 3600) (10, 3600) (5, 3600)};
\addplot+[mark=*,mark options = {scale=0.8},red] coordinates {
(60, 2.38) (50, 8.63) (40, 30.83) (30, 115.17) (20,464.33) (10, 3600) (5, 3600)};
\addplot+[mark=diamond*,mark options = {scale=0.8},black] coordinates {
(60, 0.045) (50, 0.10) (40, 0.35) (30, 1.48) (20,6.55) (10,35.23) (5,104.67) };
\end{semilogyaxis}
\end{tikzpicture}
}
\subfloat[][]{
\begin{tikzpicture}[scale=0.46]
\begin{semilogyaxis}[
					minor ytick={0.1,1,10,100,1000},
					extra y ticks={3600},
					extra y tick style={grid=major},
					extra y tick labels={\color{black}TimeOut},
					yscale = 0.92,
					xmin = 30,
					x dir = reverse,
					title={Connect},
					ylabel={CPU time(s)},
             		xlabel={minsup(\%)},
             		legend entries={\closed, \cpim, \lcmthree},
             		legend style={at={(0.53,0.55)},anchor=north west,{draw=none},font=\tiny},	
					legend columns=1
					]
\addplot+[mark=+,mark options = {scale=0.8},blue] coordinates {
(90, 12.16) (80, 59.61) (70, 146.20) (60, 300.00) (50, 600) (40, 3600) };
\addplot+[mark=*,mark options = {scale=0.8},red] coordinates {
(90, 3.69) (80, 9.73) (70, 21.82) (60, 44.18) (50, 85.30) (40, 155.63)};
\addplot+[mark=diamond*,mark options = {scale=0.8},black] coordinates {
(90, 0.040) (80, 0.071) (70, 0.078) (60, 0.130) (50, 0.169) (40, 0.417)
};
\end{semilogyaxis}
\end{tikzpicture}
}
\subfloat[][]{
\begin{tikzpicture}[scale=0.46]
\begin{semilogyaxis}[
					minor ytick={0.1,1,10,100,1000},
					extra y ticks={3600},
					extra y tick style={grid=major},
					extra y tick labels={\color{black}TimeOut},
					yscale = 0.92,
					x dir = reverse,
					title={Pumsb},
					ylabel={CPU time(s)},
             		xlabel={minsup(\%)},
             		legend entries={\closed, \cpim, \lcmthree},
             		legend style={at={(0.55,0.60)},anchor=north west,{draw=none},font=\tiny},	
					legend columns=1
					]
\addplot+[mark=+,mark options = {scale=0.8},blue] coordinates {
(95, 47.40) (90, 50.11) (85, 67.68) (80, 145.25) (75, 379.38) (70, 3600) };
\addplot+[mark=*,mark options = {scale=0.8},red] coordinates {
(95, 3600) (90, 3600) (85, 3600) (80, 3600) (75, 3600) (70, 3600) };
\addplot+[mark=diamond*,mark options = {scale=0.8},black] coordinates {
(95, 0.051) (90, 0.064) (85, 0.066) (80, 0.079) (75, 0.089) (70, 0.108) };
\end{semilogyaxis}
\end{tikzpicture}
}
\\[-4ex]
\subfloat[][]{
\begin{tikzpicture}[scale=0.46]
\begin{semilogyaxis}[
					minor ytick={0.1,1,10,100,1000},
					extra y ticks={3600},
					extra y tick style={grid=major},
					extra y tick labels={\color{black}TimeOut},
					yscale = 0.92,
					title={Retail},
					ylabel={CPU time(s)},
             		xlabel={minsup(\%)},
             		legend entries={\closed, \cpim, \lcmthree},
             		legend style={at={(0.55,0.60)},anchor=north west,{draw=none}, ,font=\tiny},	
					legend columns=1,
									xtick={1,2,...,6},
			xticklabels={10,5,1,0.5,0.1,0.05},
				xticklabel style={
anchor=base,
yshift=-\baselineskip
},
					]
\addplot+[mark=+,mark options = {scale=0.8},blue] coordinates {
(1, 129.84) (2, 129.09) (3, 129.34) (4, 129.90) (5, 163.24) (6, 286.48) 
};

\addplot+[mark=*,mark options = {scale=0.8},red] coordinates {
(1, 3600) (2, 3600) (3, 3600) (4, 3600) (5, 3600) (6, 3600) };
\addplot+[mark=diamond*,mark options = {scale=0.8},black] coordinates {
(1, 0.028) (2, 0.017) (3, 0.032) (4, 0.041) (5, 0.100) (6,0.200) };
\end{semilogyaxis}
\end{tikzpicture}
}
\subfloat[][]{
\begin{tikzpicture}[scale=0.46]
\begin{semilogyaxis}[
					minor ytick={0.1,1,10,100,1000},
					extra y ticks={3600},
					extra y tick style={grid=major},
					extra y tick labels={\color{black}TimeOut},
					yscale = 0.92,
					title={T10I4D100K},
					ylabel={CPU time(s)},
             		xlabel={minsup(\%)},
             		legend entries={\closed, \cpim, \lcmthree},
            	legend style={at={(0.52,0.25)},anchor=north west,{draw=none},font=\tiny},	
					legend columns=1,
					xtick={1,2,...,6},
					xticklabels={10,1,0.5,0.1,0.05,0.01},
				xticklabel style={
anchor=base,
yshift=-\baselineskip
},
					]
\addplot+[mark=+,mark options = {scale=0.8},blue] coordinates {
(1, 7.33) (2, 8.80) (3, 10.49) (4, 35.12) (5, 51.40) (6, 143.44) };
\addplot+[mark=*,mark options = {scale=0.8},red] coordinates {
(1, 3600) (2, 3600) (3, 3600) (4, 3600) (5, 3600) (6, 3600) };
\addplot+[mark=diamond*,mark options = {scale=0.8},black] coordinates {
(1, 0.009) (2, 0.122) (3, 0.180) (4, 0.293) (5, 0.355) (6, 1.217) };
\end{semilogyaxis}
\end{tikzpicture}
}
%
%
\subfloat[][]{
\begin{tikzpicture}[scale=0.46]
\begin{semilogyaxis}[
					minor ytick={0.1,1,10,100,1000},
					extra y ticks={3600},
					extra y tick style={grid=major},
					extra y tick labels={\color{black}TimeOut},
					yscale = 0.92,
					title={T40I10D100K},
					ylabel={CPU time(s)},
             		xlabel={minsup(\%)},
             		legend entries={\closed, \cpim, \lcmthree},
             		legend style={at={(0.52,0.25)},anchor=north west,{draw=none},font=\tiny},	
					legend columns=1,
					xtick={1,2,...,6},
					xticklabels={10,1,0.5,0.1,0.05,0.01},
				xticklabel style={
anchor=base,
yshift=-\baselineskip
}
					]
\addplot+[mark=+,mark options = {scale=0.8},blue] coordinates {
(1, 74.09) (2, 204.67) (3, 3600) (4, 3600) (5, 3600) (6, 3600) };
\addplot+[mark=*,mark options = {scale=0.8},red] coordinates {
(1, 3600) (2, 3600) (3, 3600) (4, 3600) (5, 3600) (6, 3600) };
\addplot+[mark=diamond*,mark options = {scale=0.8},black] coordinates {
(1, 0.186) (2, 2.12) (3, 4.70) (4, 61.80) (5, 344.08) (6, 3600) };
\end{semilogyaxis}
\end{tikzpicture}
}
\subfloat[][]{
\begin{tikzpicture}[scale=0.46]
\begin{semilogyaxis}[
					minor ytick={0.1,1,10,100,1000},
					xtick={5,4,...,0},
					xticklabels={0.16,0.08,0.06,0.04,0.02,0.01},
					xticklabel style={anchor=base, yshift=-\baselineskip},
					extra y ticks={3600},
					extra y tick style={grid=major},
					extra y tick labels={\color{black}TimeOut},
					yscale = 0.92,
					xtick={1,2,...,6}, 
					title={BMS-Web-View1},
					ylabel={CPU time(s)},
             		xlabel={minsup(\%)},
             		legend entries={\closed, \cpim, \lcmthree},
             		legend style={at={(0.52,0.28)},anchor=north west,	{draw=none},font=\tiny},	
					legend columns=1
					]
\addplot+[mark=+,mark options = {scale=0.8},blue] coordinates {(1, 0.86) (2, 2.30) (3, 8.05) (4, 21.05) (5 ,46.69) (6 , 3600)};

\addplot+[mark=*,mark options = {scale=0.8},red] coordinates {
(1, 3600) (2, 3600) (3, 3600) (4, 3600) (5, 3600) (6, 3600) };

\addplot[mark=diamond*,mark options = {scale=0.8},black] coordinates {
(1, 0.022) (2, 0.029) (3, 0.096) (4,0.976 ) (5 , 2.25) (6 , 6.39)};
\end{semilogyaxis}
\end{tikzpicture}
}

\caption{Computation time on datasets for different values of minimum support. \\
\hspace*{1.3cm} Table \closed vs \cpim (columns marked by "-" means that the algorithm runs out of memory).}\label{CPUtimes}
\end{figure*}

\smallskip
\noindent
{\bf Discussion.}
First we compare \closed with the most efficient CP method \cpim. CPU times of the two methods are given in
Fig.~\ref{CPUtimes}. 

Let us take the mushroom dataset, here \closed clearly outperforms \cpim. Moreover, the gains in terms of CPU time becomes significant when the {\it minsup} threshold 
decreases. For instance, with a  $minsup=0.05\%$, the \closed constraint is about 5 times faster 
than \cpim. 
If we take connect and chess datasets, \cpim performs better in terms of CPU time. 
The behavior of \closed on the two datasets can be explained by
their important density. With such datasets, the number of closed patterns is quite huge, which reduce the pruning power of the global constraint.  It is important to stress that we are testing the pruning power of our global constraint without any modification on the search part. 
On the remaining datasets, \cpim reaches  an out of memory state due to huge number of reified
constraints. For instance, if we take T40I10D100K dataset, the CP model produced by \cpim contains $|\mathcal{T}|=100\ 000$ reified constraints to express the coverage constraint, 
$2\times |\I|=2\times 1\ 000$ reified constraints to express the closure and frequency constraints. That is, the CP solver has to load in memory a CP model of $102\ 000$ reified constraint.

%
A particular observation that we can make is the fact that our global constraint is able to handle all datasets handled by the specialized algorithm \lcmthree, which is not the case of \cpim due to the size of the CP model to load. 
The result is also quite competitive comparing to \lcmthree in terms of CPU time. For instance, if we take the BMS-Web-View1 dataset with $minsup$ = 0.06\%, the extraction of all closed patterns is achieved in less than $10s$ with \closed for $0.1s$ with \lcmthree. The same observation on T10I4D100K dataset (for $minsup$ $\leq$ 0.5\%).

To complement the results, The table given in Figure~\ref{CPUtimes} reports 
a result  comparison between \closed and \cpim. For each dataset and for each value of $minsup$, we report the number of closed patterns (Col. 3), 
the number of calls to the propagate routine of {\tt Gecode} (Col. 4), the number of explored nodes of the 
search tree (Col. 5), and the memory usage (Col. 6).  
%


If we comeback to mushroom, \closed explores less nodes than \cpim and thus, we have less calls of the propagator, while on chess and connect we have the opposite. This can be explained by the fact that we do not use any particular heuristic in the search part. The second observation is that our approach is very efficient in terms of
number of propagations. For \closed, the number of propagations remains meaningless comparing  to the one using \cpim on small minsup values. This is
due to the huge number of reified constraints used in \cpim. 
As a last observation relating to memory consumption, \closed uses very low amount of memory on 
mushroom and connect, but consumes a little bit more memory on chess comparing to \cpim.

\smallskip
\noindent
At the end, comparing \closed with \lcmthree shows that  our global constraint remains competitive knowing that it just enforces domain consistency at each node without any dedicated search heuristic.


\section{Conclusion}
\label{sec-conclusion}
In this paper, we have proposed the global constraint \closed 
for Closed Frequent Pattern Mining. \closed captures the
particular semantics of the CFPM problem in order to ensure 
a polynomial pruning algorithm ensuring domain consistency.
Experiments on several known large datasets show that our
global constraint is clearly efficient and achieves scalability 
while it is a major issue for CP approaches.


\bibliographystyle{splncs03}
\bibliography{GC4IM}

\end{document}